\documentclass{article} 
\usepackage{iclr2025_conference,times}
\usepackage{graphicx}

\usepackage{amsmath,amsfonts,bm}

\def\eqref#1{equation~\ref{#1}}

\def\1{\bm{1}}

\DeclareMathAlphabet{\mathsfit}{\encodingdefault}{\sfdefault}{m}{sl}
\SetMathAlphabet{\mathsfit}{bold}{\encodingdefault}{\sfdefault}{bx}{n}

 \usepackage{multirow}
\usepackage{hyperref}
\usepackage{url}
\usepackage{booktabs}

\title{Surgical-LVLM: Learning to Adapt Large Vision-Language Model for Grounded Visual Question Answering in Robotic Surgery}

\author{%
\normalsize \textbf{Guankun Wang}$^{1,}$\thanks{Equal contribution.}\quad
\textbf{Long Bai}$^{1,}$\footnotemark[1] \quad
\textbf{Wan Jun Nah}$^{2}$ \quad
\textbf{Jie Wang}$^{1}$ \quad
\textbf{Zhaoxi Zhang}$^{1}$ \quad \\
\textbf{Zhen Chen}$^{3}$ \quad
\textbf{Jinlin Wu}$^{3}$ \quad
\textbf{Mobarakol Islam}$^{4}$ \quad
\textbf{Hongbin Liu}$^{3}$ \quad
\textbf{Hongliang Ren}$^{1,}$\thanks{Corresponding author.}
\\
\\
\normalsize $^1$ The Chinese University of Hong Kong \quad
\normalsize $^2$ Universiti Malaya \\
\normalsize $^3$ Centre for Artificial Intelligence and Robotics, HKISI-CAS \quad
\normalsize $^4$ University College London \\
\\
\small \texttt{\{gkwang, b.long\}@link.cuhk.edu.hk, hlren@ee.cuhk.edu.hk} \\
\footnotetext[2]{Corresponding authors.}
}

\iclrfinalcopy 
\begin{document}

\maketitle

\begin{abstract}
Recent advancements in surgical Visual Question Answering (VQA) and related region grounding have shown great promise for robotic and medical applications, addressing the critical need for automated methods in personalized surgical mentorship. However, existing models primarily provide simple structured answers and struggle with complex scenarios due to their limited capability in recognizing long-range dependencies and aligning multimodal information. In this paper, we introduce Surgical-LVLM, a novel personalized large vision-language model tailored for complex surgical scenarios. Utilizing the pre-trained large vision-language model and specialized Visual Perception LoRA (VP-LoRA) blocks, our model excels in understanding complex visual-language tasks within surgical contexts. In addressing the visual grounding task, we propose the Token-Interaction (TIT) module, which strengthens the interaction between the grounding module and the language responses of the Large Visual Language Model (LVLM) after projecting them into the latent space. We demonstrate the effectiveness of Surgical-LVLM on several benchmarks, including EndoVis-17-VQLA, EndoVis-18-VQLA, and a newly introduced EndoVis Conversations dataset, which sets new performance standards. Our work contributes to advancing the field of automated surgical mentorship by providing a context-aware solution.
\end{abstract}

\section{Introduction}
\label{sec:1}
The progress achieved in surgical vision question answering (VQA)~\citep{seenivasan2022surgical} and related region grounding has attracted many researchers since these techniques demonstrate great potential for applications in robotics and healthcare. To integrate the above techniques and achieve them simultaneously, \cite{bai2023surgical} pioneered the surgical visual question localized-answering (VQLA), which predicts answers with localized results through the input text and images. Current approaches in this field predominantly rely on the expertise of expert surgeons, often failing to acknowledge the challenges posed by real-world constraints, such as the limited availability of expert surgeons. Consequently, the demand for auto methods for personalized surgical mentorship stands as a critical, yet challenging issue. Despite their effectiveness in providing simple answers, these methods generally have difficulties when faced with challenging answer-location scenarios.

Building upon the groundwork of elementary Visual Question Answering (VQA) models, ~\citep{seenivasan2023surgicalgpt,yuan2023advancing,bai2023cat,bai2023revisiting} innovatively delineates a paradigm for the community in the study of Surgical Visual Question Localized-Answering utilizing deep learning methodologies.
However, these methods are mainly subject to two common limitations. Firstly, these methods have difficulty establishing accurate long-range dependencies when dealing with complex scenarios, due to occlusion and distortion, which may lead to the ignorance of relevant visual details that are significant to surgical environments. Additionally, they often overlook the alignment of multiple modalities, resulting in a mismatch between language and visual grounding. Considering these issues above, we ask: \textit{Can we personalize a large vision-language model to automatically ground VQA in surgical scenarios in a simple and efficient manner?}

\begin{figure*}[thpb]
    \centering
    \includegraphics[width=0.96\linewidth, trim=0 120 210 0]{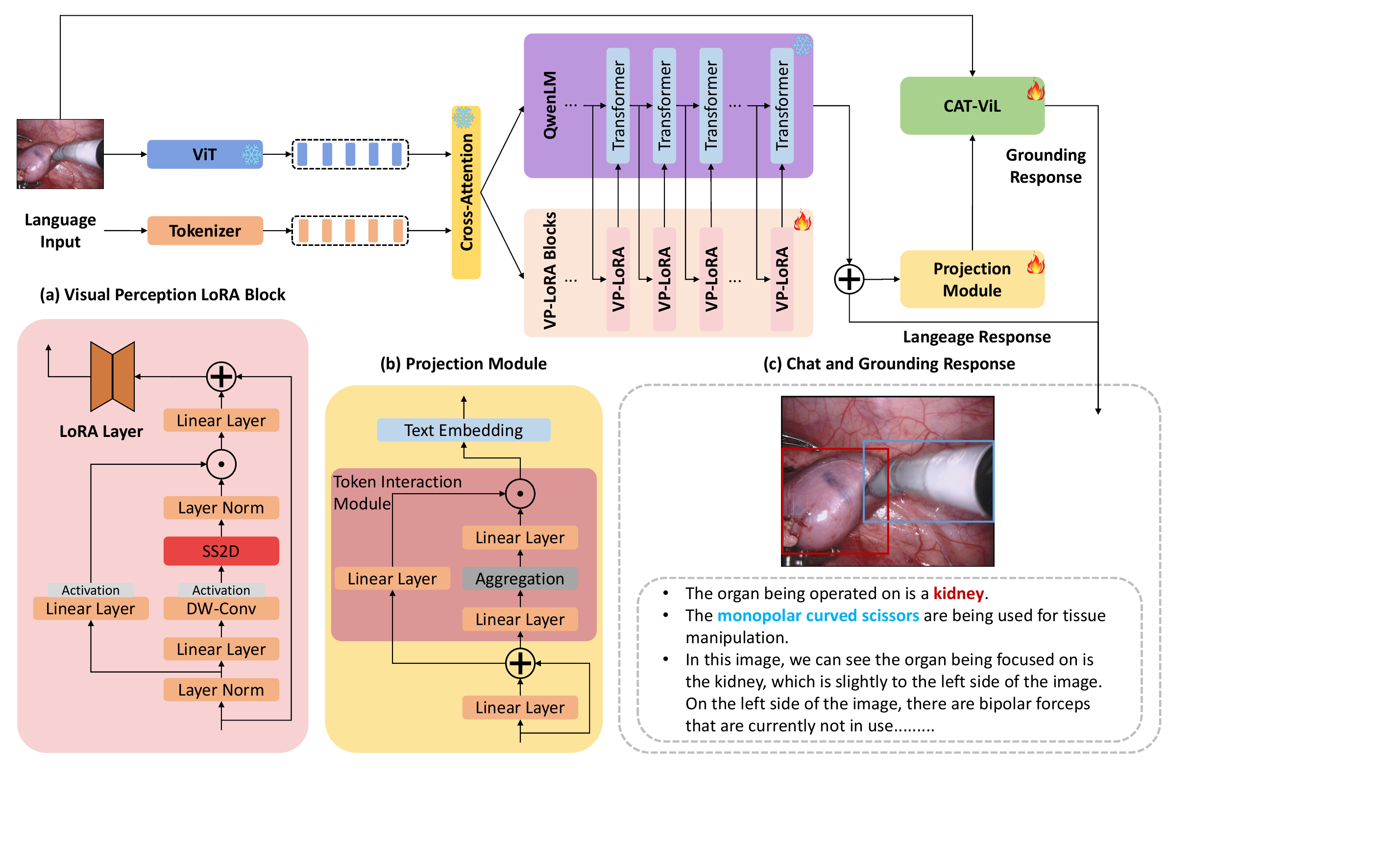}
    \caption{The flowchart of the proposed framework. It includes a pre-trained ViT, tokenizer, the Qwen-VL model, projection module, and the CAT-ViL embedding module.}
    \label{fig:overview}
\end{figure*}

With the rise of Large Language Models (LLMs)~\citep{achiam2023gpt}, Large Vision-Language Models (LVLMs) have emerged as a new method for understanding the interplay between visual and textual information~\citep{liu2024llava,zhang2023llavagrounding,you2023ferret}. However, existing large models still struggle to grasp domain-specific semantic information in subdomains~\citep{li2024llavamed}, making a surgical-focused LVLM a current necessity. To this end, we present Surgical-LVLM, a personalized surgical large vision-language model for highly specialized surgical scenarios. As shown in Fig.~\ref{fig:overview}(c), our method efficiently customizes LVLM faced with occlusive medical samples. Specifically, we first obtain the describable visual-language tokens by a pre-trained vision transformer (ViT) and Qwen-VL~\citep{bai2023qwen}. Consequently, we utilize a cross-attention layer stack of specially designed Visual Perception LoRA (VP-LoRA) blocks to unleash attention's feature caption power with one core technique. Specifically, we integrate Visual State Space (VSS) blocks at each Low-Rank Adaptation layer, enabling the effective propagation of global contextual information and allowing Qwen-VL to better adapt to specialized surgical scenarios.

With the aforementioned designs, Surgical-LVLM exhibits favorable grounding performance for hard samples in a variety of complex scenes. Nevertheless, there might be occasional failure cases, specifically the misalignment problem that the language model may describe some correct, yet subimportant content. 
Such ambiguity challenges Surgical-LVLM's grounding capacity in determining the appropriate object to localize as output since both the important and sub-important parts can be regarded as correct answers. To alleviate this issue, we further propose an alignment variant of our approach. 
In detail, we project the tokens generated from Qwen-VL to latent space and take a token interaction (TIT) module to enable text tokens to emphasize the important vision information. By such efficient projection-based training, Surgical-LVLM selectively focuses on the relevant features or words that are most important for the VQLA task. Our contributions are as follows:

\vspace{-1.5mm}
\begin{itemize}
    \item We introduce Surgical-LVLM, a personalized foundational model tailored for VQA and grounding for intricate surgical contexts. Additionally, we further incorporate VP-LoRA to facilitate efficient long-range representation learning in complex surgical environments.
    \item We present the Token-Interaction Module, designed to effectively project and align language responses from LLMs with the visual grounding module, enhancing the synergy between textual and visual data.
    \item We have extensively validated the effectiveness of our proposed method on multiple benchmarks, including the public EndoVis 2017 and 2018 datasets, and our novel EndoVis Conversations dataset. The results significantly underscore improvements in accurate surgical visual-language understanding and grounding, and establish state-of-the-art performance on these datasets. 
\end{itemize}

\section{Related Work}
\label{sec:2}
\subsection{Large Visual-Language Model}
With the significant progress of LLMs such as GPT-4~\citep{achiam2023gpt} and its open-source alternatives like LLaMA~\citep{touvron2023llama} and Vicuna~\citep{chiang2023vicuna}, researchers have been increasingly motivated to exploit their integration of powerful language abilities with vision~\citep{zhang2023llava}. Recently, the performance of the general large multimodal models has been continuously refreshing the expectations since the scaling up of the pretraining data~\citep{bai2023qwen, dai2024instructblip} and instruction-following data~\citep{dai2024instructblip, li2023mimic, wang2024copesd}. However, most models can only excel in general domains and easily encounter challenges in specific scenarios, such as complex surgical scenarios. In addition, while much of the current work has a good performance of language responses based on images and user instructions, it is difficult to provide alignment to related image regions (visual grounding)  simultaneously. It is noted that CogVLM-Grounding~\citep{wang2023cogvlm} and MiniGPT-v2~\citep{chen2023minigpt} have a limitation in the generation of concise captions during grounding tasks, attributable to its primary training on caption data from datasets such as Flickr30K~\citep{plummer2015flickr30k}. Conversely, BuboGPT~\citep{zhao2023bubogpt} preserves conversational competency utilizing an external grounding model for grounding purposes, although the language encoder in the grounding model constrains the chat capability. These methods face challenges in achieving proficiency in both chat and grounding domains. 

\subsection{VQLA Models in Medical Domain}
Medical VQLA has emerged as a pivotal area in the medical domain, using the integration of textual questions and images to support clinical diagnostics and treatment. However, the unique characteristics inherent in medical data necessitate specialized expertise for accurate interpretation, thereby introducing additional complexities in developing effective medical VQLA models~\citep{liu2023q2atransformer, bai2025surgical}. Recent advancements in medical VQLA research have been focused on addressing these inherent challenges. Surgical-VQLA~\citep{bai2023surgical} incorporates a gated vision-language embedding approach to fuse multimodal features from visual and textual sources. CAT-ViL~\citep{bai2023cat} is subsequently proposed as a co-attention gated embedding module, allowing for instructive interaction between text and visual embeddings. These pioneering endeavors have contributed to several specialized improvements for medical scenarios and achieved superior representation learning. However, due to parameter limitations and the lack of specialized data, these models can just answer fixed-format questions and are unable to analyze comprehensive information about surgical scenarios to meet clinical needs. Therefore, the introduction of large language models~\citep{liu2023improved, touvron2023llama, zheng2024judging} into the field of surgical-VQLA is necessitated by their powerful capabilities in text generation and comprehension.

\section{Methodology}
\label{sec:3}

\subsection{Preliminaries}
\label{sec:3.1}

\subsubsection{Qwen-VL} Qwen-VL is a Large Vision-Language Mode designed to perceive and understand both texts and images~\citep{bai2023qwen}. The model architecture consists of Qwen-7B as the language model, a Vision Transformer (ViT) as the visual encoder, and a vision-language adapter to compress the image features. During training, the model is trained in a three-stage pipeline: pretraining with image-text pairs, multi-task pretraining with interleaved vision-language data, and supervised fine-tuning. This training process enables Qwen-VL to interact with users, perceive input images, and generate appropriate responses based on user intent.

\subsubsection{Surgical-VQLA} Surgical-VQLA indicates models for Visual Question Localized-Answering in robotic surgery~\citep{bai2023surgical}, which uses a combination of VisualBERT~\citep{li2019visualbert} and a Gated Vision-Language Embedding (GVLE) technique for effective fusion of visual and text features. The GVLE is designed to build input patches for the Language Vision Transformer and enables localized answering by incorporating a detection head parallel to the prediction head of the LViT. Besides, CAT-ViL~\citep{bai2023cat} further introduces the Co-Attention Gated Vision-Language (CAT-ViL) embedding module, which fuses multimodal features more effectively.

\subsection{Surgical-LVLM}

\label{sec:3.2}
We propose Surgical-LVLM to perform complex scene comprehension as well as logical reasoning while performing VQLA tasks in surgical scenarios. As shown in Fig.~\ref{fig:overview}, the framework consists of finetuned Qwen-VL (ViT~\citep{dosovitskiy2020image}, tokenizer, QwenLM and VP-LoRA blocks), a projection module and the grounding module (CAT-ViL).

\subsubsection{Visual Perception LoRA}
\label{sec:3.2.1}
Due to problems such as occlusion and distortion, traditional Transformers have difficulties in establishing accurate long-range dependencies when dealing with images in complex surgical scenarios. Therefore, we design Visual Perception LoRA, which enhances the visual perception capabilities of Qwen-VL by integrating Visual State Space (VSS) blocks~\citep{liu2024vmamba} at each Low-Rank Adaptation (LoRA) layer~\citep{hu2021LoRA} during finetuning. 

As shown in Fig.~\ref{fig:overview} (a), the input first passes through an initial normalization layer. Then, the output is divided into two parallel information branches: the first branch contains linear transformation and activation, while the second branch incorporates linear transformation, depthwise separable convolution, and subsequent activation, followed by selective feature extraction through the 2D-Selective-Scan (SS2D) module. The SS2D is able to capture different features by expanding them and conducting feature extraction~\citep{ruan2024vm}. Finally, the linear layer is utilized to integrate the merged features of the two branches, which are then combined via the residual connection to generate the VP-LoRA output following the LoRA layer. With $X_{in}$ and $X_{out}$ representing the input and output features, the equation of the VP-LoRA block is as follows: 
\begin{equation}
    X_{out} = \mathcal{F}_{\mathrm{LoRA}}[\mathcal{F}_{\mathrm{VSS}}(X_{in})]
\end{equation}

The incorporation of VSS blocks into each LoRA layer of the Qwen-VL model facilitates the efficient learning of global contextual information. At each layer, the VSS blocks capture the dependencies between visual features and allow the model to refine its understanding of the input images. Besides, the VP-LoRA enables the model to focus on relevant visual details, understand complex visual scenes, and generate more accurate and contextually relevant responses.

\begin{figure*}[thpb]
    \centering
    \includegraphics[width=0.96\linewidth, trim=0 200 80 0]{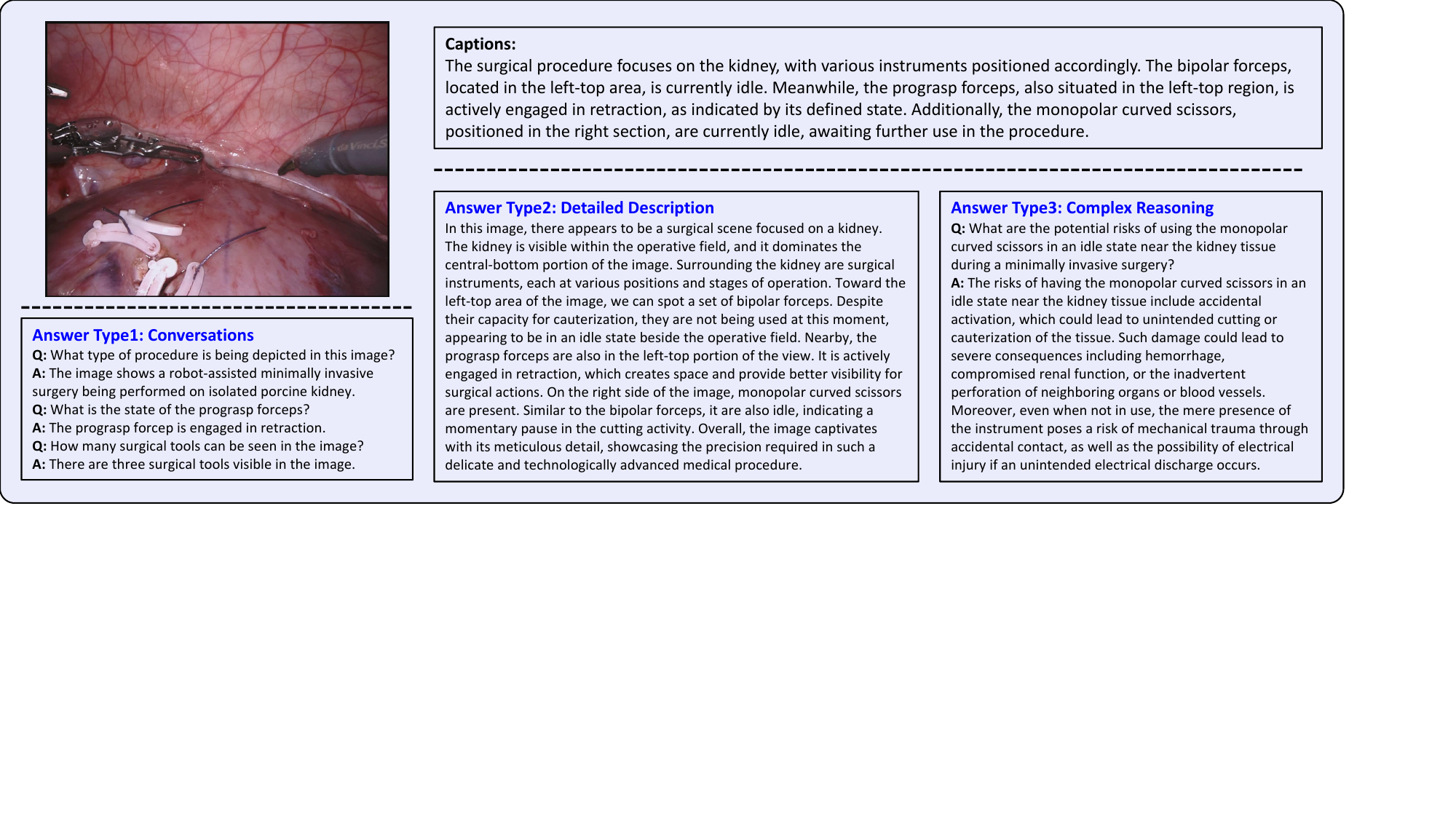}
    \caption{A sample of Instruction Tuning data generation. Contexts such as captions and images used to prompt the GPT are shown above the dashes, and three types of responses are shown below the dashes. We use images as prompts for GPT, which can help GPT comprehend the questions better and generate appropriate responses.}
    \label{fig:ins}
\end{figure*}
\subsubsection{Multimodal Alignment for Visual Grounding}
\label{sec:3.2.2}
Currently, many LVLMs have separate chat and grounding functions when conducting the multimodel task. However, these models tend to perform poorly when both chatting and grounding are required due to the misalignment problem -- the language model may describe some correct, yet subimportant content, which will mislead grounding model predictions. To tackle this problem, we propose a novel projection module to align the language response generated from Qwen-VL with the input of CAT-ViL. We design the projection module to enable the language response to support the prediction of related grounding responses. 

In the alignment projection module, the language response is first processed through a linear layer and then combined with a residual connection. The combined features will be passed through the Token InTeraction (TIT) module~\citep{yang2022focal} to further refine and emphasize the important information in the features. There are two parallel flows in the TIT module. The first flow contains two linear layers with an Aggregation block between them. With $x_{i}$ and $y_{i}$ representing the input and output features, the equations of the Aggregation block are as follows:
\begin{equation}
    \boldsymbol{y}_{i}=\mathcal{F}_{inter}\left[\mathcal{A}(i, \mathbf{X}), \boldsymbol{x}_{i}\right]
\end{equation}
where the contextual features for each location $i$ are aggregated using $\mathcal{A}$ first, and then the query is computed via interaction $\mathcal{F}_{inter}$ to generate $\boldsymbol{y}_{i}$. $\mathcal{A}$ denotes the Aggregation operation. The second branch is just linear transformation. Finally, the merged features of these two flows will be output to the CAT-ViL model for text embedding. The TIT module selectively focuses on the relevant features or words that are most important for the VQLA task.

Through multimodal alignment with our projection module, we combine the capabilities of Qwen-VL and CAT-ViL to continuously generate language and grounding responses, thus achieving a fusion of chat and grounding abilities. This integration allows for a comprehensive understanding of the surgical scene and facilitates effective communication between the system and the user.

\subsection{Instruction Tuning}
\label{sec:3.3}
In order to enhance the performance of LVLMs' capability for surgical scenarios understanding and logical reasoning, it is essential to create a dataset with high quality. For this purpose, we have established a dataset comprising question-answer pairs based on the images and annotations from the EndoVis 2017/2018 VQLA datasets. These pairs have been formatted according to the Qwen-VL structure. Guided by the methodology~\citep{liu2024visual}, we have formulated questions that span a spectrum of complexity, ranging from easy conversations, detailed descriptions, and complex reasoning as illustrated in Figure~\ref{fig:ins}. These instruction-tuning QA pairs have been synthesized utilizing the GPT-4, which is renowned for its exemplary linguistic proficiency. The detailed statistical information on this dataset can be found in Sec~\ref{sec:4.1}.

\subsection{Training Strategy}
\label{sec:3.4}
Our training strategy for Surgical-LVLM consists of two stages: instruction turning for LVLM and multimodel alignment for the grounding module.

\textbf{Vision-Language Instruction Fine-tuning:}
\label{sec:3.4.1}
In the first phase, we utilize the extensive dataset comprising surgical question-answer pairs alongside associated images to fine-tune the Qwen-VL model augmented with VP-LoRA. This stage exclusively modifies the parameters of adapter layers while keeping the original large language model layers frozen. Through instruction fine-tuning, the Qwen-VL model acquires comprehension of the semantic intricacies and linguistic structures specific to the surgical domain from both textual and visual encoding features. Thus, the model can generate meaningful representations, adeptly capturing the nuances and intricacies inherent in surgical terminology and context, thus establishing a robust foundation for effective vision-language grounding.

\textbf{Visual Grounding Alignment:}
\label{sec:3.4.2}
In the second stage of training, the fusion of chat and grounding capabilities is enacted by integrating the Qwen-VL and visual grounding module. Utilizing the pre-trained Qwen-VL model, language features extracted from its outputs are employed alongside corresponding surgical images to comprehensively optimize the parameters of CAT-ViL embedding. This optimization is geared towards enhancing the coherence between language responses and specific visual regions in surgical images. To achieve this alignment, we use the co-attention embedding fusion to effectively combine the multi-modal features. Through this process, CAT-ViL acquires the capacity to accurately ground textual descriptions to pertinent regions in the surgical images by learning the associations between Qwen-VL's output features and visual representations.

Overall, through the integration of the instruction fine-turned Qwen-VL and the trained CAT-ViL model, our proposed two-stage training strategy culminates in the synergistic enhancement of chat and grounding functionalities. This approach maximizes the utilization of Qwen-VL's logical reasoning VQA abilities and the language grounding capabilities of CAT-ViL, thereby achieving an integrated fusion of these components to fully harness their potential.

\section{Experiments}
\label{sec:4}

\subsection{Dataset}
\label{sec:4.1}

\textbf{EndoVis Conversations Dataset} 
is a novel dataset generated from EndoVis-18-VQLA and EndoVis-17-VQLA Datasets. We designed questions encompassing three distinct types using GPT-4. Statistically, the training and test sets include 19,020 and 2,151 image-QA pairs, respectively. The image setup for the training and test sets is also referenced to the above two datasets.

\textbf{EndoVis-18-VQLA Dataset} 
comprises 14 robotic surgery videos obtained from the MICCAI Endoscopic Vision Challenge~\citep{allan2020endovis18}. Within this dataset, VQLA annotations are incorporated, integrating the question-answer pairs derived from~\citep{seenivasan2022global} alongside bounding box annotations sourced from~\citep{islam2020learning}. These QA pairs encompass a wide spectrum of single-word responses that encapsulate information regarding organs, surgical instruments, and their interrelations. We adhere to the data partitioning strategy delineated in~\citep{seenivasan2022surgical} for the training and test datasets.

\textbf{EndoVis-17-VQLA Dataset} is emanated from the MICCAI Endoscopic Vision Challenge 2017~\citep{allan2019endovis17}. Utilizing this dataset as an external validation benchmark, we employ it to demonstrate the generalization efficacy of our proposed model across diverse surgical scenarios.

\subsection{Implementation Details}
\label{sec:4.2}
 We utilize the mainstream GPT-4 Score in the comparison experiments against BLIP-2~\citep{li2023blip}, miniGPT-4~\citep{zhu2023minigpt}, LLaVA-1.5~\citep{liu2023improved} and LLaVA-Med~\citep{li2024llava} on the EndoVis Conversations Dataset. Besides, for the EndoVis-18-VQLA and EndoVis-17-VQLA Datasets, the baseline models for our quantitative experimental comparison refer to~\citep{bai2023cat}. The evaluation metrics are accuracy, f-score, and mean intersection over union (mIoU)~\citep{rezatofighi2019generalized}. All experiments are implemented by Python PyTorch framework, and conducted on a server with NVIDIA A100 GPUs. The epoch, batch size, and learning rate are set to 20, 16, and $1 \times 10^{-5}$, respectively.

\begin{table}[h]
\renewcommand{\arraystretch}{1.1}
\caption{Ablation study on EndoVis Conversation dataset. ‘VP-LoRA’ denotes ‘Visual Perception LoRA’ and ‘IF’ denotes ‘Instruction Fine-tuning’.}
\label{tab:abl1}
\centering
\resizebox{0.7\textwidth}{!}{
\begin{tabular}{c|cc|cc}
\toprule[1pt]
\multirow{2}{*}{Model}         & \multirow{2}{*}{VP-LoRA} & \multirow{2}{*}{IF} & \multicolumn{1}{c|}{EndoVis-18-C} & EndoVis-17-C \\ \cline{4-5} 
                               &                      &                           & \multicolumn{2}{c}{GPT-4 Score}                        \\ \hline
\multirow{3}{*}{Surgical-LVLM} & $\times$                     & $\times$                          & \multicolumn{1}{c|}{72.48}           & 69.62           \\\cline{2-5}
                               & \checkmark                    & $\times$                          & \multicolumn{1}{c|}{74.61}           & 70.20            \\\cline{2-5}
                               & $\times$                     & \checkmark                         & \multicolumn{1}{c|}{88.53}           & 81.23           \\\cline{2-5}
                               & \checkmark                    & \checkmark                         & \multicolumn{1}{c|}{\textbf{90.68}}           & \textbf{83.24}           \\ \toprule[1pt]
\end{tabular}}
\end{table}

\begin{table}[h]
\renewcommand{\arraystretch}{1.1}
\caption{Comparison experiments of our Surgical-LVLM against LVLMs without fine-tuning on EndoVis Conversations Dataset.}
\label{tab:fine}
\centering
\resizebox{0.7\textwidth}{!}{
\begin{tabular}{c|cc}
\toprule[1pt]
                                     & \multicolumn{1}{c|}{EndoVis-18-C} & EndoVis-17-C \\ \cline{2-3} 
\multirow{-2}{*}{Model}              & \multicolumn{2}{c}{GPT-4 Score}                        \\ \hline
BLIP-2~\citep{li2023blip}                               & \multicolumn{1}{c|}{65.21}           & 60.43           \\ \hline
miniGPT-4~\citep{zhu2023minigpt}                            & \multicolumn{1}{c|}{70.79}           & 66.51           \\ \hline
LLaVA-1.5-7B~\citep{liu2023improved}                         & \multicolumn{1}{c|}{80.20}            & 76.11           \\ \hline
LLaVA-1.5-13B~\citep{liu2023improved}                        & \multicolumn{1}{c|}{81.29}           & 77.59           \\ \hline
LLaVA-Med~\citep{li2024llava}                            & \multicolumn{1}{c|}{83.54}           & 80.63           \\ \hline
Surgical-LVLM & \multicolumn{1}{c|}{\textbf{90.68}}           & \textbf{83.24}           \\ \toprule[1pt]
\end{tabular}}
\end{table}

\begin{table}[h]
\renewcommand{\arraystretch}{1.1}
\caption{Quantitative evaluation experiments on EndoVis-18 and EndoVis-17 datasets.}
\label{tab:main}
\centering
\resizebox{\textwidth}{!}{
\begin{tabular}{c|ccc|ccc}
\toprule[1pt]
\multirow{2}{*}{Model} & \multicolumn{3}{c|}{EndoVis-18-VQLA} & \multicolumn{3}{c}{EndoVis-17-VQLA} \\ \cline{2-7} 
                       & Acc        & F-Score     & mIoU      & Acc        & F-Score    & mIoU      \\ \hline
VisualBERT~\citep{li2019visualbert}             & 0.6268     & 0.3329      & 0.7391    & 0.4005     & 0.3381     & 0.7073    \\ \hline
VisualBERT R~\citep{seenivasan2022surgical}      & 0.6301     & 0.339       & 0.7352    & 0.419      & 0.337      & 0.7137    \\ \hline
MCAN~\citep{ben2017mutan}                   & 0.6285     & 0.3338      & 0.7526    & 0.4137     & 0.2932     & 0.7029    \\ \hline
VQA-DeiT~\citep{touvron2021training}               & 0.6104     & 0.3156      & 0.7341    & 0.3797     & 0.2858     & 0.6909    \\ \hline
MUTAN~\citep{ben2017mutan}                  & 0.6283     & 0.3395      & 0.7639    & 0.4242     & 0.3482     & 0.7218    \\ \hline
MFH~\citep{yu2018beyond}                    & 0.6283     & 0.3254      & 0.7592    & 0.4103     & 0.35       & 0.7216    \\ \hline
BlockTucker~\citep{ben2019block}            & 0.6201     & 0.3286      & 0.7653    & 0.4221     & 0.3515     & 0.7288    \\ \hline
CAT-ViL DeiT~\citep{bai2023cat}           & 0.6452     & 0.3321      & 0.7705    & 0.4491     & \textbf{0.3622}     & 0.7322    \\ \hline
GVLE-LViT~\citep{bai2023surgical}              & 0.6659     & \textbf{0.3614}      & 0.7625    & \textbf{0.4576}     & 0.2489     & 0.7275    \\ \hline
Surgical-LVLM (Ours)       & \textbf{0.6947}     & 0.3325      & \textbf{0.8416}    & 0.4068     & 0.3412     & \textbf{0.7825}    \\ \toprule[1pt]
\end{tabular}}
\end{table}

\begin{table}[h]
\renewcommand{\arraystretch}{1.1}
\caption{Ablation study on different modified modules. ‘VP-LoRA’ denotes ‘Visual Perception LoRA’ and ‘MA’ denotes ‘Multimodal Alignment’.}
\label{tab:abl2}
\centering
\resizebox{\textwidth}{!}{
\begin{tabular}{c|cc|ccc|ccc}
\toprule[1pt]
\multirow{2}{*}{Model}         & \multirow{2}{*}{VP-LoRA} & \multirow{2}{*}{MA} & \multicolumn{3}{c|}{EndoVis-18-VQLA} & \multicolumn{3}{c}{EndoVis-17-VQLA} \\ \cline{4-9} 
                               &                      &                     & Acc        & F-Score     & mIoU      & Acc        & F-Score    & mIoU      \\ \hline
\multirow{4}{*}{Surgical-LVLM} & $\times$                     & $\times$                    & 0.6878     & 0.3211      & 0.8303    & 0.4054     & 0.3386     & 0.7689    \\ \cline{2-9} 
                               & \checkmark                    & $\times$                    & 0.6947     & 0.3325      & 0.8322    & 0.4068     & 0.3412     & 0.7719    \\ \cline{2-9} 
                               & $\times$                     & \checkmark                   & 0.6878     & 0.3211      & 0.8336    & 0.4054     & 0.3386     & 0.7781    \\ \cline{2-9} 
                               & \checkmark                    & \checkmark                   & \textbf{0.6947}     & \textbf{0.3325}      & \textbf{0.8416}    & \textbf{0.4068}     & \textbf{0.3412}     & \textbf{0.7825}    \\ \toprule[1pt]
\end{tabular}}
\end{table}

\subsection{Results}
\label{sec:4.3}

We first evaluate the effectiveness of our proposed methods on the EndoVis Conversation dataset in Table~\ref{tab:abl1}. The Surgical-LVLM framework is evaluated based on whether VP-LoRA is introduced and whether it undergoes Instruction Fine-tuning. We can find that Instruction Fine-tuning is crucial for logical reasoning and comprehensive responses in surgical scenarios for the large language model. This is demonstrated by the huge gap in the GPT-4 Score. Therefore, the generalized large language models are not yet fully applicable in specialized scenarios such as surgery. Moreover, the introduction of VP-LoRA brings an effective enhancement to the model's language response, proving its effectiveness in accurate response generation.

Next, we compared the performance of our Surgical-LVLM framework against excellent LVLMs without Instruction Fine-tuning in surgical scenarios, which is shown in Table~\ref{tab:fine}. Significantly, Surgical-LVLM has demonstrated remarkable performance with the highest GPT-4 Score recorded at 90.68 on the EndoVis-18-VQLA dataset and 83.24 on the EndoVis-17-VQLA dataset, surpassing all competing models evaluated on the EndoVis Conversations Dataset. This notable margin of improvement further underscores the imperative for instruction tuning within LVLMs frameworks when applied to specific domains.

Quantitative evaluation in Table~\ref{tab:main} presents that our Surgical-LVLM outperforms other state-of-the-art models in traditional Surgical-VQLA tasks. On the EndoVis-18 dataset, our Surgical-LVLM model outperforms the others in terms of Accuracy and mIoU but has a lower F-Score. This indicates it is better at identifying the correct answers and detecting the images but might struggle slightly with balancing precision and recall. Its high mIoU suggests strong grounding capabilities. On the EndoVis-17 dataset, the performance of all models decreased due to suffering from domain migration. Despite the overall lower scores compared to EndoVis-18, the Surgical-LVLM still has the highest mIoU, confirming its robustness. However, other models, such as CAT-ViL DeiT and BlockTucker, have slightly better Accuracy and F-scores, hinting at potential areas of improvement for the Surgical-LVLM in terms of Accuracy and F-scores. Overall, the Surgical-LVLM model with Qwen large language model as well as multi-model alignment of CAT-ViL, which has excellent reasoning ability as well as strong inference-based grounding capability, as evidenced by mIoU metrics that are far superior to those of other state-of-the-art models.

Furthermore, an ablation study is conducted to observe the effects of the designed modules in EndoVis-18-VQLA and EndoVis-17-VQLA datasets. Surgical-LVLM is categorized by the inclusion of VP-LoRA and Multimodal Alignment and the results are shown in Table~\ref{tab:abl2}. Incorporating VP-LoRA consistently enhances performance across both datasets and metrics. While MA shows improvement, its impact is focused on detection performance. Models combining both VP-LoRA and MA demonstrate the highest performance, indicating synergistic effects. Overall, these findings emphasize the importance of VP-LoRA in the propagation of global contextual information for accurate response generation, with complementary benefits when combined with MA for grounding.

\subsection{Discussion and Limitations}
\label{sec:4.4}
The quantitative evaluation demonstrates the superiority of our proposed Surgical-LVLM framework over other state-of-the-art models. The ablation analysis on the EndoVis Conversation dataset elucidates the pivotal role of VP-LoRA and Instruction Fine-tuning in augmenting language response performance across datasets and metrics. Notably, comparing Surgical-LVLM with other LVLMs further reinforces the necessity of Instruction Fine-tuning for domain-specific applications. Besides, Surgical-LVLM achieves higher accuracy and mIoU on the EndoVis-18 dataset. This achievement is primarily attributed to the VP-LoRA with the supplementary integration of Multimodal Alignment yielding further enhancements in grounding capabilities. 

However, performance degradation across all models on the EndoVis-17 dataset underscores the challenge of maintaining robustness across different domains. While the Surgical-LVLM outperformed on various metrics, its slightly lower F-Score on the EndoVis-18 dataset suggests room for improvement in balancing precision and recall. In addition, further research could explore VP-LoRA's optimal integration and potential limitations, although its introduction enhances response accuracy. Finally, the deployment of LVLMs in real surgical scenarios is still limited by safety concerns necessitating a rigorous evaluation of their reliability in complex surgical environments to mitigate potential risks.

\section{Conclusions}
\label{sec:5}
In this paper, we present Surgical-LVLM, a personalized vision-language model specifically designed for efficient grounded visual question answering in robotic surgery. The Surgical-LVLM uses the specifically designed VP-LoRA to enable the model to understand complex visual-language surgical tasks. Additionally, we introduce a novel token interaction module to facilitate the critical language information emphasis. Our experimental results, conducted on multiple benchmarks, manifest the superiority of our proposed approach. However, deploying LVLMs in real-world surgical scenarios remains a long way off due to the need to thoroughly review the reliability and decision-making process of LVLMs in dynamic and unpredictable surgical environments. Therefore, future works shall highlight the need for ethical considerations, and robust safety mechanisms to realize the potential of foundation models in surgical applications. 

\subsubsection*{Acknowledgments}
This work was supported by Hong Kong RGC CRF C4026-21GF, GRF 14203323, GRF 14216022, GRF 14211420, NSFC/RGC Joint Research Scheme N\_CUHK420/22, Shenzhen-Hong Kong-Macau Technology Research Programme (Type C) STIC Grant 202108233000303.

\bibliography{iclr2025_conference}
\bibliographystyle{iclr2025_conference}

\end{document}